\documentclass[graybox, envcountchap]{svmult}

\usepackage{mathptmx}        
\usepackage{helvet}          
\usepackage{courier}         

\usepackage{makeidx}         
\usepackage{graphicx}        
\usepackage{multicol}        
\usepackage[bottom]{footmisc}

\makeindex             

\begin{document}

\frontmatter

%
%
%

\begin{dedication}
Use the template \emph{dedic.tex} together with the Springer document class SVMono for monograph-type books or SVMult for contributed volumes to style a quotation or a dedication\index{dedication} at the very beginning of your book in the Springer layout
\end{dedication}

%
%

\foreword

Use the template \textit{foreword.tex} together with the Springer document class SVMono (monograph-type books) or SVMult (edited books) to style your foreword\index{foreword} in the Springer layout. 

The foreword covers introductory remarks preceding the text of a book that are written by a \textit{person other than the author or editor} of the book. If applicable, the foreword precedes the preface which is written by the author or editor of the book.

\vspace{\baselineskip}
\begin{flushright}\noindent
Place, month year\hfill {\it Firstname  Surname}\\
\end{flushright}

%
%

\preface

Use the template \emph{preface.tex} together with the Springer document class SVMono (monograph-type books) or SVMult (edited books) to style your preface in the Springer layout.

A preface\index{preface} is a book's preliminary statement, usually written by the \textit{author or editor} of a work, which states its origin, scope, purpose, plan, and intended audience, and which sometimes includes afterthoughts and acknowledgments of assistance. 

When written by a person other than the author, it is called a foreword. The preface or foreword is distinct from the introduction, which deals with the subject of the work.

Customarily \textit{acknowledgments} are included as last part of the preface.

\vspace{\baselineskip}
\begin{flushright}\noindent
Place(s),\hfill {\it Firstname  Surname}\\
month year\hfill {\it Firstname  Surname}\\
\end{flushright}

%
%

\extrachap{Acknowledgements}

Use the template \emph{acknow.tex} together with the Springer document class SVMono (monograph-type books) or SVMult (edited books) if you prefer to set your acknowledgement section as a separate chapter instead of including it as last part of your preface.

\tableofcontents
%
%
%
\contributors

\begin{thecontriblist}
Firstname Surname
\at ABC Institute, 123 Prime Street, Daisy Town, NA 01234, USA, \email{smith@smith.edu}
\and
Firstname Surname
\at XYZ Institute, Technical University, Albert-Schweitzer-Str. 34, 1000 Berlin, Germany, \email{meier@tu.edu}
\end{thecontriblist}
%
%

\extrachap{Acronyms}

Use the template \emph{acronym.tex} together with the Springer document class SVMono (monograph-type books) or SVMult (edited books) to style your list(s) of abbreviations or symbols in the Springer layout.

Lists of abbreviations\index{acronyms, list of}, symbols\index{symbols, list of} and the like are easily formatted with the help of the Springer-enhanced \verb|description| environment.

\begin{description}[CABR]
\item[ABC]{Spelled-out abbreviation and definition}
\item[BABI]{Spelled-out abbreviation and definition}
\item[CABR]{Spelled-out abbreviation and definition}
\end{description}

\mainmatter
%
%
%

\begin{partbacktext}
\part{Part Title}
\noindent Use the template \emph{part.tex} together with the Springer document class SVMono (monograph-type books) or SVMult (edited books) to style your part title page and, if desired, a short introductory text (maximum one page) on its verso page in the Springer layout.

\end{partbacktext}

\title*{Modeling Multiple User Interests using Hierarchical Knowledge for Conversational Recommender System}
\titlerunning{Modeling Multiple User Interests using Hierarchical Knowledge for CRS}
\author{Yuka Okuda, Katsuhito Sudoh, Seitaro Shinagawa, and Satoshi Nakamura}
\institute{Yuka Okuda \at Nara Institute of Science and Technology, Ikoma, Nara, Japan, \email{okuda.yuka.ou0@is.naist.jp}}
%
%
\maketitle


\abstract{
A conversational recommender system (CRS) is a practical application for item recommendation through natural language conversation. Such a system estimates user interests for appropriate personalized recommendations. Users sometimes have various interests in different categories or genres, but existing studies assume a unique user interest that can be covered by closely related items. In this work, we propose to model such multiple user interests in CRS. We investigated its effects in experiments using the ReDial dataset and found that the proposed method can recommend a wider variety of items than that of the baseline CR-Walker.
}






\section{Introduction}
\label{sec:1}
         \begin{figure}[h]
         \label{fig:one_pointproblem}
        \centering \includegraphics[width=1.0\linewidth]{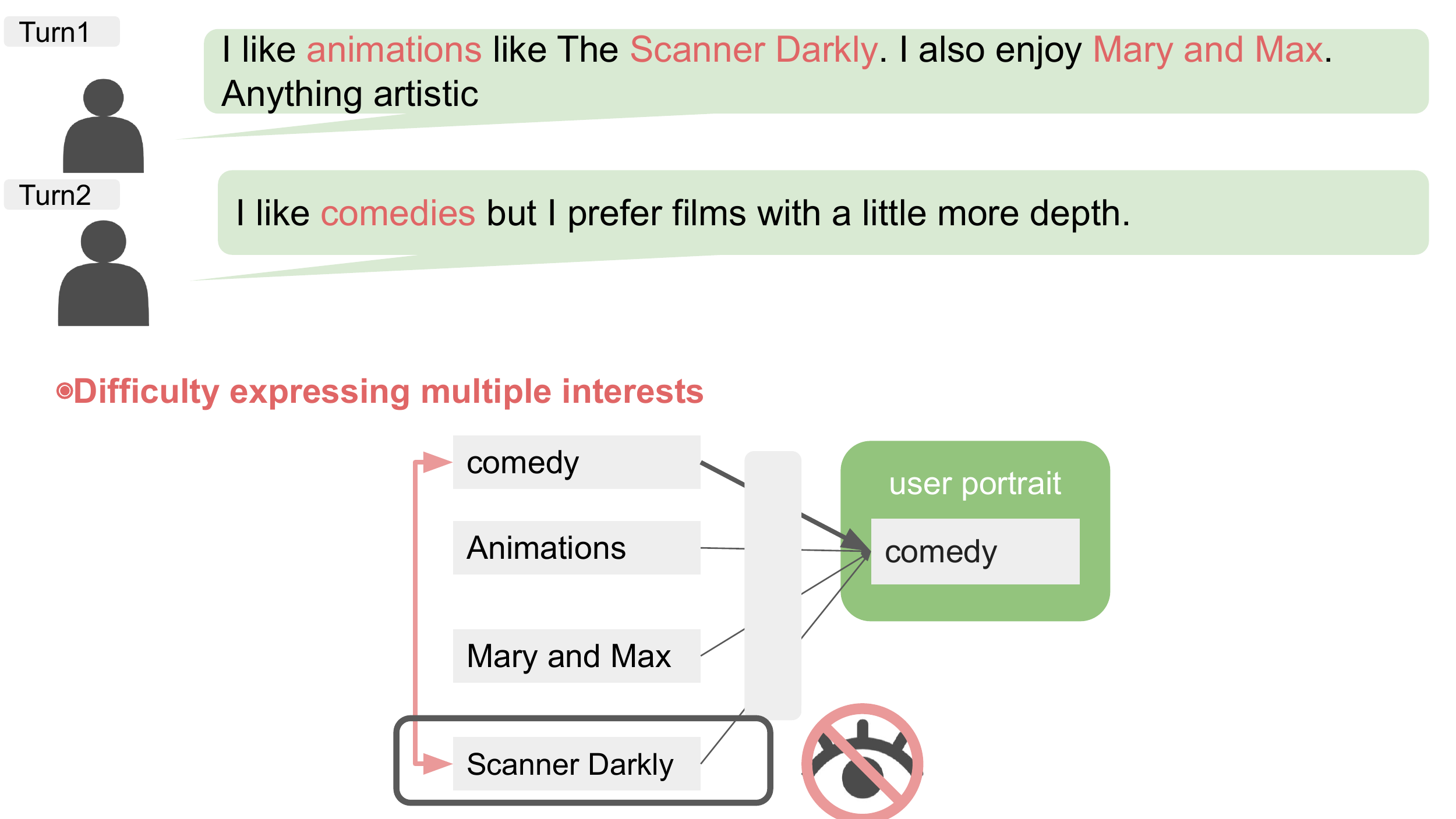}
        \caption{An illustration of a difficult case for existing systems to capture multiple interests of users: two example utterances are shown at the top, red text in user utterances indicates entities, entities extracted from user utterances are represented by gray boxes and arranged vertically, and the green frame represents the user portrait.}
        \end{figure}
    Recommender System is an attractive field of research and development for many commercial applications.
    A typical recommender system recommends items to users using collaborative filtering \cite{Beel2016,Collaborative_Filtering} based on a large amount of accumulated data from other users' choices.
    A major drawback of this approach is the so-called \emph{cold start problem} \cite{Gope2017ASO};
    when a target user has no history in order to identify his/her interests and preferences for the recommendation.
    Interaction with users can mitigate this problem by iteratively updating their interests and preferences.
    Natural language conversation is a promising way for interaction between users and recommender systems, especially for new under-experienced users.
    Conversational Recommender System (CRS) \cite{Li2018,KGSF} is a variant of such a recommender system.
    
    CRS recommends items to users according to their \emph{user portrait} through conversation.
    The user portrait is a representation of user interests used for the recommendation \cite{ma2021crwalker}.
    Existing CRS studies \cite{KGSF,ma2021crwalker} represent a user portrait using a user-dependent embedding vector and use it to choose appropriate items for recommendation.
    However, such a representation leads to a limited variety of item recommendations because a portrait vector corresponds to one point in the embedding space. 
    In cold start situations, users present their interests step by step through conversation so their interests are not always specific enough to be represented in such a way.
    Furthermore, in the case of users having wide interests ranging over different categories or genres, this representation would fail to capture their multiple interests.
    

    Fig.~\ref{fig:one_pointproblem} shows an example in a movie domain.
    A user mentions four entities (two genres and two titles) in two utterances, then the CRS estimates the user's portrait from the utterances.
    These genres and titles are important clues to estimate the user portrait, but the representation by a single portrait embedding vector calculated as a weighted sum of the embeddings mixes these different interests.
    As a result, the CRS fails to capture the users' interests ranging over different genres such as animation, comedy, and science fiction (derived from Scanner Darkly) and makes recommendations focusing on just one genre of comedy.
    If the user asks the system to recommend more items for wider coverage, it should recommend similar items related to comedy.
    
    
    In this paper, we tackle this problem and propose a method that models a user portrait with multiple interests considering different granularity by the use of hierarchical knowledge of items and genres.
    Using the hierarchical knowledge, we can keep multiple interests in different branches with appropriate granularity, such as a genre of comedy and a specific item of Scanner Darkly.
    In experiments using Redial dataset, 
    the proposed method recommended a wider variety of items than the baseline using CR-Walker.

\section{Related Work}
\label{sec:2}
There have been two major approaches to natural language-based recommendation systems. 
One is called Interactive Information Retrieval (IIR), which asks users questions in natural language. 
Zhang et al. \cite{Towards_Conversational_Search} proposed an information retrieval system that asks explicit questions to fill the slots describing user interests.
The other is CRS which mimics natural language dialogues for the recommendation by humans. 
It has advantages in its naturalness and sophisticated dialogue strategies to explore and narrow down users' interests.
This work is based on the latter approach and aims to leverage these advantages to obtain various user interests in the recommendation task.

\label{sec:related_crs}
      \begin{figure}[t]
        \centering
        \includegraphics[width=0.8\linewidth]{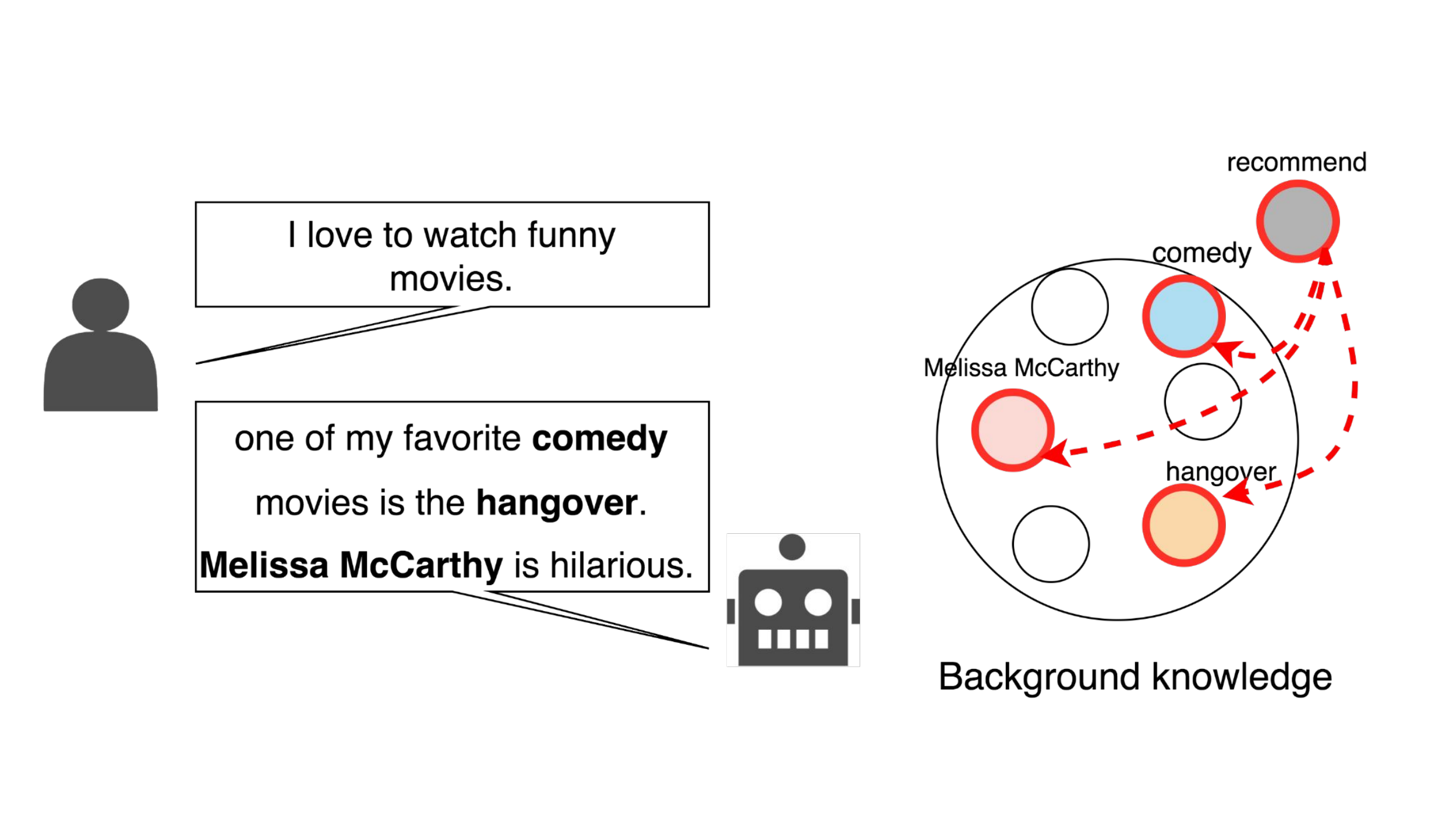}
        \vspace{-7mm}
        \caption{CRS using background knowledge.}
        \label{fig:system_over} 
      \end{figure}

CRS has attracted many studies in recent years along with the advance in the field of natural language dialog systems and chatbots.
One important problem in CRS is incorporating background knowledge to make accurate recommendations and generate appropriate system utterances.
The background knowledge is knowledge of what knowledge the system uses in utterances.
Zhou et al. \cite{KGSF} and Ma et al. \cite{ma2021crwalker} focus on the relationship between dialogue strategies and background knowledge for these purposes.
In these existing studies, the background knowledge is estimated using user portraits, utterance contexts, relationships between knowledge, and so on.
CR-Walker \cite{ma2021crwalker} uses a reasoning tree to obtain accurate background knowledge. 
However, these studies do not model a user portrait to capture multiple user interests.

With respect to the problem of multiple user interests, Qi et al. \cite{qi-etal-2021-hierec} proposed a recommendation system that considers multiple user interests based on click rates in Web browsing.
However, their approach cannot be applied to CRS because their system is based on a large-scale click rate data.

In this work, we focus on the problem of multiple user interests in CRS and propose a method to model user portraits that are capable of capturing multiple interests.

\section{CR-Walker: a conventional CRS model with user portrait}

\label{sec:3}
\begin{figure*}[t]
    \centering
    \includegraphics[width=1.0\linewidth]{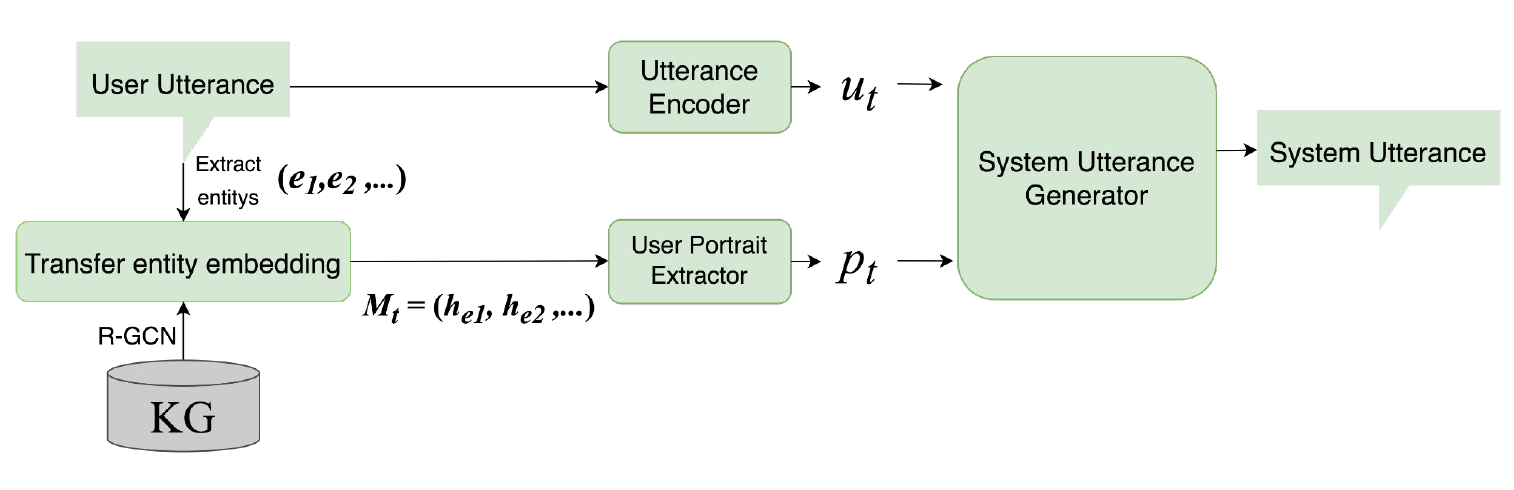}
    \caption{Overview of CRS.}
    \label{fig:CR-walker}
\end{figure*}

Before we present our proposed method, we describe CR-Walker\cite{ma2021crwalker} in this section\footnote{Due to the space limitation, we omit some details that would not be required in this paper. Please refer to the literature for its detailed formulation.}.
CR-Walker is a conventional CRS method using user portraits. 
It consists of three modules: an utterance encoder, a user portrait extractor, and a system utterance generator using a reasoning tree.
Its overall architecture is illustrated in Fig.~\ref{fig:CR-walker}.

\subsection{Utterance encoding}
Suppose a user and a CRS take turns in their conversation and let ${x}_{t}$ and ${y}_{t}$ be user and systems utterances, respectively.
The utterance encoder captures a contextual representation of the conversation between a user and the system.
User and system utterances are converted into embeddings using BERT \cite{devlin-etal-2019-bert} and then converted into contextual embeddings using LSTM \cite{sak2014long}.
The contextual embedding of the conversation until the user utterance at the time step $t$, ${u}_{t}$, is denoted as:
\begin{align}
\label{form:Utterance_Encoder}
\boldsymbol{u}_{t} = \operatorname{LSTM_{\text{hidden}}}\left(\boldsymbol{u}_{t-1}, \operatorname{BERT}\left(\left[y_{t-1} ; x_{t}\right]\right)\right),
\end{align}
where the subscript ``hidden'' indicates the vector ${u}_{t}$ is the hidden vector of the LSTM at time $t$ and not its output vector.
The input to the LSTM is the concatenation of ${y}_{t-1}$ and ${x}_{t}$.
\subsection{User portrait extraction}
\label{subsec:3-2}

The user portrait extractor derives a user portrait vector ${p}_{t}$ from the user utterances ${x}_{1}, \ldots, {x}_{t}$ and a knowledge graph.
It does not use whole user utterances but focuses only on mentioned entities.
Here, we assume that the knowledge graph represents relations among a set of named entities $\mathcal{N}$.

First, the user portrait extractor finds named entities $E_{t} = e_{1}, \ldots, e_{N_{t}}$ from the user utterances using named entity recognition,
where $N_{t}$ is the number of mentioned entities in ${x}_{1}, \ldots, {x}_{t}$.
The entities are converted into the corresponding entity embeddings ${M}_{t} =\{h_{1}, \ldots, h_{N_{t}} \}$.
Each entity embedding $h_{i}$ is calculated through a $L$-layered network, and its intermediate embedding at $l$-th layer is denoted as:
\begin{align}
\boldsymbol{h}_{i}^{(l)}=\sigma \left( \left(\sum_{r \in \mathcal{R}} \sum_{e' \in \mathcal{N}_{e_i}^{r}} \frac{1}{\left|\mathcal{N}_{e_i}^{r}\right|} \boldsymbol{W}_{r}^{(l-1)} \boldsymbol{h}_{e'}^{(l-1)} \right) + \boldsymbol{W}_{0}^{(l-1)} \boldsymbol{h}_{i}^{(l-1)}\right),
\end{align}
where $\mathcal{N}_{e_i}^{r}$ is the set of neighboring entities of $e_{i}$ under the relation $r$ in the knowledge graph,
$\boldsymbol{h}_{e'}^{(l-1)}$ is the embedding vector for the entiry $e'$ in $\mathcal{N}_{e_i}^{r}$ at $(l-1)$-th layer,
$\boldsymbol{W}_{r}^{(l)} \in \mathbb{R}^{d \times d}$ and $\boldsymbol{W}_{0}^{(l)} \in \mathbb{R}^{d \times d}$ are learnable matrices for integrating relationship-specific information from the neighboring and current entities.
$h_{*}^{0}$ is derived from an embedding matrix $W_{\text{emb}} \in \mathbb{R}^{d \times \mathcal{N}}$;
this means that the user portrait extractor model covers all the named entities in $\mathcal{N}$ regardless of their appearance in the training data.
The knowledge graph is extracted from DBpedia and then represented using R-GCN \cite{R-GCNschlichtkrull2018modeling}.
Then, the user portrait $\boldsymbol{p}_{t} \in \mathbb{R}^{d}$ is derived through attention onto $\boldsymbol{M}_{t}$ as follows:
\begin{align}
\label{eq:userportrate}
\boldsymbol{p}_{t} &=\boldsymbol{\alpha}_{t} * \boldsymbol{M}_{t}, \\
\boldsymbol{\alpha}_{t} &=\operatorname{softmax}\left(\boldsymbol{w}_{p} \cdot \tanh \left(\boldsymbol{W}_{p} \boldsymbol{M}_{t}\right)\right),
\end{align}
where $\boldsymbol{w}_{p}$ is the weight for each entity embedding and $\boldsymbol{W}_{p}$ is the weight matrix for the self-attention.
\subsection{System utterance generation}
\label{subsec:3-3}
According to the user portrait extracted by the procedure above, the system generates responses to ask more questions and give recommendations to the user.
The generation process consists of: (1) making a reasoning tree to determine what will be mentioned in the response and (2) generating a response using the reasoning tree and a language generation model.
        
\subsubsection{Reasoning Tree}
\label{subsubsec:3-3-1}
\begin{figure*}[h]
    \centering
    \includegraphics[width=0.8\linewidth]{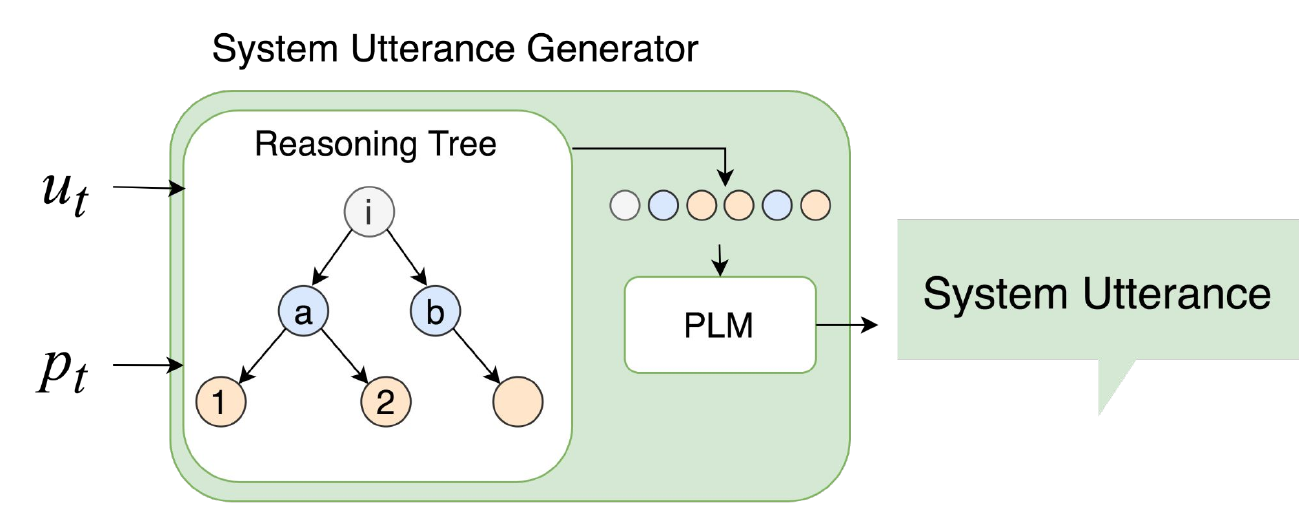}
    \caption{System Utterance Generator.}
    \label{fig:Ut_gen}
\end{figure*}



A reasoning tree is a tree-structured graph as shown in Fig.~\ref{fig:Ut_gen}.
CR-Walker uses two types of nodes for dialogue acts and knowledge graph entities, both represented by their embedding vectors derived from $u_{t}$ and $p_{t}$.

The root node stores the dialog act embedding $i_{t}$ denoted as follows:
\begin{align}
i_{t} = \boldsymbol{W}_{int}^{2} \operatorname{ReLU}\left(\boldsymbol{W}_{int}^{1} \boldsymbol{u}_{t}\right),
\end{align}
where $\boldsymbol{W}_{int}^{1}$ and $\boldsymbol{W}_{int}^{2}$ are weight matrices.
There are three dialog acts: \emph{Recommend}, \emph{Query}, and \emph{Chat}, and the system changes its behaviors according to it as described later.

Vectors associated with the descendant nodes come from embeddings of knowledge graph entities but are also influenced by their ancestor nodes.
A context embedding $c_{t}$ is defined for each descendant node as follows\footnote{We omit a superscript specifying the node for simplicity.}:
\begin{align}
        \boldsymbol{c}_{t} &=\gamma_{t}  \boldsymbol{u}_{t}+\left(1-\gamma_{t}\right) \boldsymbol{p}_{t},\\
        \gamma_{t} &= \begin{cases}
             \sigma\left(\boldsymbol{w}_{1} \cdot\left[\boldsymbol{u}_{t} ; \boldsymbol{p}_{t} ; \boldsymbol{i}_{t}\right]\right), & \text{(if the parent is the root)} \\
             \sigma\left(\boldsymbol{w}_{2} \cdot\left[\boldsymbol{u}_{t} ; \boldsymbol{p}_{t} ; \boldsymbol{i}_{t} ; \boldsymbol{h}_{e_\text{parent}}\right]\right), & \text{(otherwise)}
            \end{cases}
\end{align}
where $w_1$ and $w_2$ indicate weight vectors, $\boldsymbol{h}_{e_\text{parent}}$ is the embedding of the entity associated with the parent node, and $[a ; b]$ represents concatenation of vectors.

Suppose we are going to append a child node associated with an entity $e^{\prime}$ to the reasoning tree.
We define a score $\hat{s}_{e^{\prime}}$ using context embeddings over its ancestor nodes as follows:

\begin{align}
\hat{s}_{e^{\prime}}=\left(\boldsymbol{h}_{e^{\prime}} \cdot \left(\boldsymbol{c}_{t} + \boldsymbol{c}_{t}^\text{parent}\right) \right),
\end{align}

where $\boldsymbol{c}_{t}$ is the context vector of the node to which we append the child node, and $\boldsymbol{c}_{t}^\text{parent}$ is that of its parent node.
Note that $\boldsymbol{c}_{t}^\text{parent}$ is calculated recursively up to the root node.
Based on the score, all of the entities that satisfy the function $\operatorname{WALK}(e)=\left\{e^{\prime} \mid \hat{s}_{e^{\prime}}>\tau \right\}$ form new nodes in the reasoning tree\footnote{$\operatorname{WALK}(e)$ returns the top-1 entity if all the entities scored below $\tau$.}.
The threshold hyperparameter $\tau$ controls the choice of entities included in the reasoning tree.
The node appending step is repeated by $N$ times to induce an $(N+1)$-layer reasoning tree.
CR-Walker induces a three-layer reasoning tree whose nodes represent different types of entities according to dialog acts, as shown in Table~\ref{table:hop-rure}.
The dialog act \emph{Recommend} lets the system recommend items to the user.
\emph{Query} and \emph{Chat} let the system ask questions to a user and say something about mentioned entities, respectively.


\begin{table}[t]
 \caption{Types of entities stored in the reasoning tree for different dialog acts.}
 \label{table:hop-rure}
  \centering
 \begin{tabular}{|l|l|l|}
\hline Dialog act (root) & Middle layer & Leaf layer \\
\hline Recommend & Attributes of mentioned items & Candidate items  \\
\hline Query & Generic classes & Attributes \\
\hline Chat & Mentioned entities & All entities \\
\hline
 \end{tabular}
\end{table}

\subsubsection{Utterance generation}
\label{subsubsec:3-3-2}
Based on the reasoning tree, the system generates utterances.
The reasoning tree is converted into a sequence  and then used as an input prompt for a pre-trained language model to predict a system utterance that follows the input prompt.

\section{Proposed method}
\label{sec:4}
%
We propose a method that extends the \textit{User Portrait Extractor} of CR-Walker to capture a wide range of user interests.
%
The proposed method uses a hierarchical structure of the entity knowledge and considers multiple abstract item classes (i.e., genres) explicitly to calculate a user portrait vector.
In this work, we assume the item hierarchy of the entity knowledge: genres and titles in the case of the movie domain.
Our motivation is to capture various user interests ranging over different genres.
User portraits induced by CR-Walker are based on entity embeddings and focus on a limited number of entities suggested through attention (Eqs. (3)-(4)).
In contrast, the proposed method aims to capture user interests in different genres using explicit genre-level user portraits in the reasoning tree induction.

\subsection{Hierarchical item knowledge}
\label{subub-sec:4-4-2}
    \setlength\textfloatsep{10pt}
    \begin{figure}[h]
            \centering
            \includegraphics[width=1.0\linewidth]{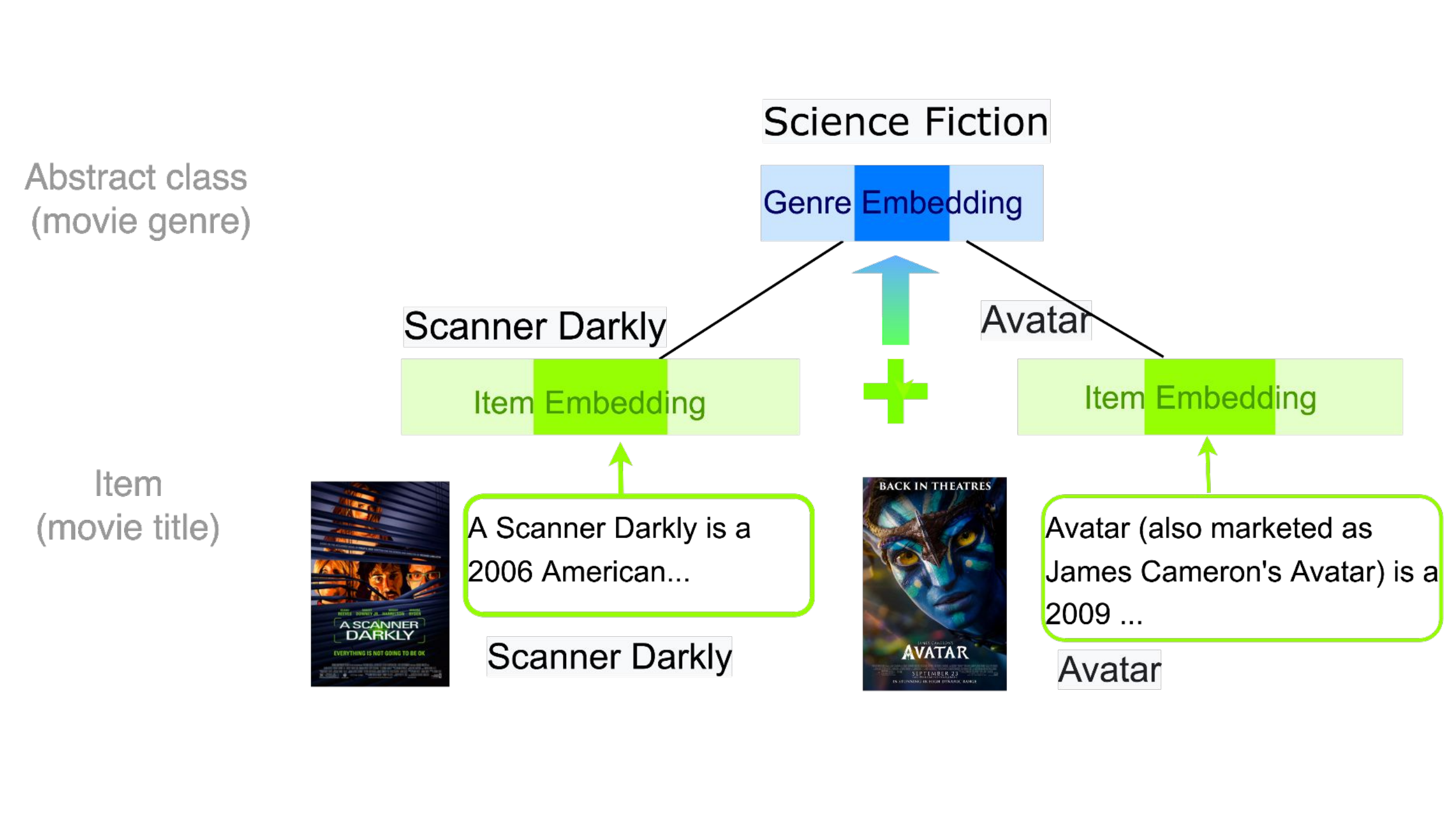}
            \caption{Constructing hierarchical knowledge.}
            \label{fig:hierarical_graph}
    \end{figure}


We assume that item knowledge in the target domain is available with the information on item classes.
We use the knowledge to construct a hierarchical structure
as shown in Fig.~\ref{fig:hierarical_graph},
which has item nodes (e.g., movie titles) and abstract class nodes (e.g., movie genres).
Here, we assume each item is associated with only one abstract class for simplicity.

We assign embedding vectors to the nodes in the hierarchical knowledge.
The embedding vector for each item node is given simply by the average of word embeddings for nouns and adjectives in its item description: the abstract in the case of movies. 
Each abstract class node is given by the average of the embeddings of associated item nodes. 
The embedding model is a pre-trained one such as word2vec.
For example in Fig.~ \ref{fig:hierarical_graph}, the embedding of the genre node \emph{Science Fiction} is the average of the embeddings of the item nodes \emph{Scanner Darkly} and \emph{Avatar}.

\subsection{User portrait extraction considering multiple interests}
We calculate the similarity between a user utterance and each node embedding in the hierarchical knowledge as the cosine between these vectors.
The vector representation of the user utterances is given in the same way as for the item node.
The node score is updated by user utterances in the dialog;
the cosine similarity is accumulated as the score of each node, ${S}_{t}^{H}$.
Using these node scores, we extract a user portrait vector ${p}_t^{H}$ and use it for the middle layer of the reasoning tree to constrain the recommendation with the user's interests in the abstract classes.
%
%
We choose two attribute entities with the highest scores, ${e}^{H}_{1}$ and ${e}^{H}_{2}$, to consider multiple user interests ranging over different attributes.
Their corresponding entity embeddings ${h}^{H}_{1}$ and ${h}^{H}_{2}$ are used to calculate the user portrait ${p}_t^{H}$ as follows:
    \begin{align}
    \boldsymbol{p}_{t}^{H} &=\boldsymbol{\alpha}_{t}^{H} * \boldsymbol{M}^{H}_{t}, \\ 
    \boldsymbol{\alpha}_{t}^{H} &=\operatorname{softmax}\left(\boldsymbol{w}_{p} \cdot \tanh \left(\boldsymbol{W}_{p} \boldsymbol{M}^{H}_{t}\right)\right), \\
    {M}^{H}_{t}&=({h}^{H}_{1},{h}^{H}_{2}).
    \end{align}
        
Note that we use the same $\boldsymbol{w}_{p}$ and $\boldsymbol{W}_{p}$ as CR-Walker.
And we use the user portrait ${p}_t^{H}$ only for the attribute-level reasoning and still use the original user portrait ${p}_t$ in the item-level reasoning.
    
\section{Experiments}
\label{sec:5}
We investigate the performance of the proposed method through the following experiments to compare it with the baseline of CR-Walker.

\subsection{Setup}
For the experiments, we used ReDial \cite{Li2018}, a public conversation recommendation dataset in the movie domain.
Each recommendation conversation was performed between two crowd workers; one played the role of a recommender and the other was a seeker. 
Recommendation utterances by the recommender were annotated with their correct recommendation items (movie titles).
In each conversation, at least four different movies are mentioned.

As the baseline, we used the authors' implementation of CR-Walker\footnote{\url{https://github.com/truthless11/CR-Walker}} with some modifications:
(1) Entities used for $M_t$ to induce a user portrait in Eq. (3) were limited to movie genres for consistent comparisons with the proposed method;
(2) When the dialog act is \emph{Recommend}, the system induces a reasoning tree by choosing top-1 entity among entities from movie attributes such as genre, year, actor, etc. in the middle layer and top-2 movie titles in the leaf layer\footnote{We observed the threshold-based entity selection described in 3.3.1 results in top-1 entities in most cases, so this modification would not affect the performance of CR-Walker seriously.}.

Resources used for the proposed method were prepared as follows.
The hierarchical knowledge of movie genres and titles came from the CR-Walker implementation, which was originally extracted from DBpedia.
The abstract description for each movie title was extracted from the corresponding DBpedia entry and Wikipedia article;
we extracted it from DBpedia entries when it was available, otherwise, we apply a simple matching pattern of ``\emph{[movie name] is \ldots}'' to extract the abstract description.
For the word embeddings used in the proposed method, we used a pre-trained word2vec model of \textit{GoogleNews-vectors-negative30word2vec}\footnote{\url{https://code.google.com/archive/p/word2vec}}and Natural Language Toolkit\footnote{\url{https://www.nltk.org/}}.


\subsection{Results}
\label{subsec:6-1}
\begin{table}[t]
  \begin{minipage}[]{.4\textwidth}
 \caption{Recall@K (R@K) and coverage (Cov.) of the recommendations.}
    \begin{center}
        \begin{tabular}{|l|l|l|l|l|}
          \hline
          Model & R@1 & R@10 & R@50 & Cov. \\
          \hline \hline
          CR-Walker & 3.33 & 14.6  & 30.5 & 17.4\\ 
          Proposed & 3.23 & 14.5 & 30.2 & \textbf{21.1}\\ 
          \hline
      \end{tabular}
    \vspace{-2mm}
    \end{center}
    \label{table:Recall-Coverage}
  \end{minipage}
  \qquad
  \begin{minipage}[]{.55\textwidth}
    \caption{BLEU, distinct (dist)-1/2/3, and F1 of utterance generation results.}
    \begin{center}
      \begin{tabular}{|l|l|l|l|l|l|}
          \hline
          Model & BLEU & dist-1 & dist-2& dist-3& F1 \\
          \hline \hline
          CR-Walker & 7.62 & 5.5 & 23.0 & 39.1 & 15.64\\ 
          Proposed & 7.64 & \textbf{6.1} & \textbf{24.3} & \textbf{40.4} &15.63 \\ 
          \hline
      \end{tabular}
      \vspace{-2mm}
    \end{center}
    \label{table:BLEU-Dist-F1}
  \end{minipage}
\end{table}

\begin{table}[t]
  \centering
  \small
  \caption{Example of the recommendation and utterance generation by the baseline CR-Walker and the proposed method. Middle layer entities (attributes) and leaf layer entities (items) are derived from a reasoning tree.}
    \label{table:multi-exp}
  \begin{tabular}{l|l|p{4.0cm}|p{1.5cm}|p{1.3cm}|p{2.3cm}}
    \hline
    Turn
    & Model
    & Utterance
    & Dialog act
    & Middle layer
    & Leaf layer \\ \hline\hline
    \multirow{3}{*}{1}
    &User& \multicolumn{4}{p{9cm}}{i love to watch \textbf{funny movies}}\\\cline{2-6}
    &CR-Walker & one of my favorite comedy movies is the hangover. melissa mccarthy is hilarious. i also really like bridesmaids & Recommend & Comedy & (1) The Hangover (2) Bridesmaids\\ \cline{2-6}
    &Proposed  & one of my favorite comedy movies is the hangover. melissa mccarthy is hilarious. i also really like bridesmaids & Recommend & Comedy & (1) The Hangover (2) Bridesmaids\\  \hline
    \multirow{3}{*}{2}
    &User& \multicolumn{4}{p{9cm}}{yes, i love adam sandler}\\\cline{2-6}
    &CR-Walker & have you seen click or the waterboy?  & Recommend & American films  & (1)~Click (2)~The Waterboy\\ \cline{2-6}
    &Proposed  & have you seen click or the waterboy?  & Recommend & Comedy & (1)~Click (2)~The Waterboy\\  \hline
    \multirow{3}{*}{3}
    &User& \multicolumn{4}{p{9cm}}{just watched the mask last night! there is a new one Son of the Mask. i like some \textbf{scary movie}}\\\cline{2-6}
    &CR-Walker & have you seen tommy boy or ace ventura: pet detective?  & Recommend & \textbf{Comedy} & (1)~Tommy Boy (2)~Ace Ventura: Pet Detective\\ \cline{2-6}
    &Proposed  & have you seen tucker \& dale vs. evil? or leprechaun?  & Recommend & \textbf{Horror}  & (1)~Tucker and Dale vs. Evil (2)~Leprechaun\\  \hline

  \end{tabular}
  \label{tab:sample2}
\end{table}

First, we measured the performance of the recommendation using Recall@K and coverage.
Recall@K is the recall of K-best retrieval results, which is widely used in information retrieval.
The coverage measures the variety of item recommendations by the fraction of the number of unique recommended items (for a certain test set) and the size of the item set.
Table~\ref{table:Recall-Coverage} shows the results.
The proposed method did not outperform CR-Walker in Recall@K but achieved better coverage.
This suggests the proposed method covers a wider range of items in its recommendation than CR-Walker while it also causes to degrade the accuracy of the recommended items.

Then, we measured the performance of the utterance generation using BLEU \cite{papineni-etal-2002-bleu}, distinct-N \cite{li-etal-2016-diversity}, and F1.
Table~\ref{table:BLEU-Dist-F1} shows the results.
The proposed method achieved slightly higher distinct scores than CR-Walker while the BLEU and F1 were almost the same.
This suggests the proposed method generates diverse utterances due to the change in the recommendation strategy and coincides with the finding from the previous results.


            

\subsection{Analysis}

            
    
\subsubsection{Recommendation and utterance generation}
\label{subsubsec:5-3-1}
We analysed the recommendation and utterance generation results by CR-Walker and the proposed method.
Table~\ref{table:multi-exp} shows an example.
middle layer (attribute) and leaf layer (items) in a reasoning tree, which are also used to constrain the utterance generation.
The user indicates the interest in funny (\emph{comedy}) movies in the first utterance, and the both systems behave the same.
They work similarly in the next turn.
In the third turn, the user mentioned the interest in scary (\emph{horror}) movies.
While CR-Walker still traps in the genre of comedy, the proposed method successfully captures the user's interest in horror and recommends movies with comedy and horror aspects.

\subsubsection{Transition of user interests across different genres}
\begin{landscape}
        \begin{figure*}[t]
                \includegraphics[width=1.0\linewidth]{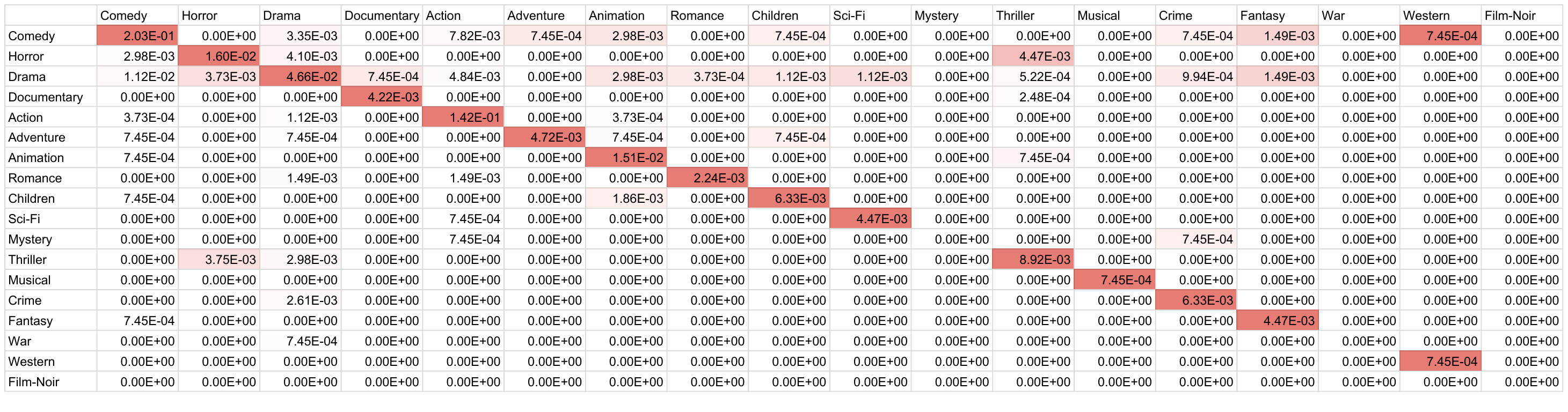}
                \vspace{-7mm}
                \caption{Transition probability of CR-Walker.}
                \label{fig:trans_prob_b}
                \vspace{10mm}
                \includegraphics[width=1.0\linewidth]{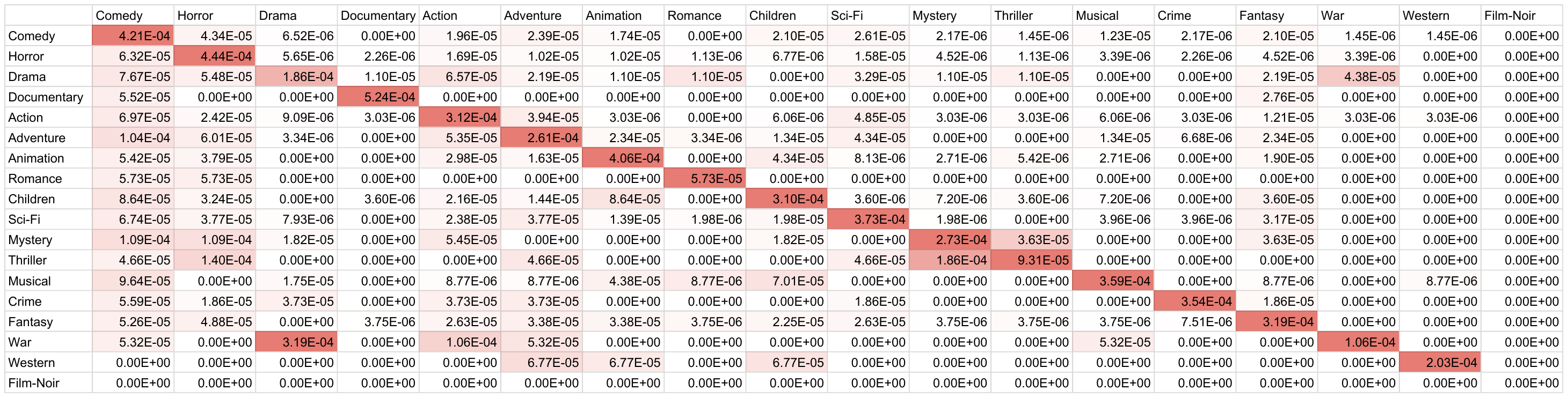}
                \vspace{-7mm}
                \caption{Transition probability of the proposed method.}
                \label{fig:trans_prob_p}
        \end{figure*}
\end{landscape}

We investigated the transition of user interests across different genres that  happened in the experiments.
Tables~\ref{fig:trans_prob_b} and \ref{fig:trans_prob_p} represent transition probabilities across middle layer (attribute) within one conversation by CR-Walker and  the proposed method, respectively.
Each cell in the tables are colored according to the corresponding transltion probability.
Their difference clearly indicates the proposed method yields more top-1 attribute transitions than CR-Walker.
This suggests the proposed method can adapt the possible change of a user portrait due to multiple user interests.

\section{Conclusions}
In this work, we proposed a method to model multiple user interests for CRS.
The proposed method uses hierarchical knowledge of abstract classes and items and incorporates multiple abstract classes explicitly into user portraits used for recommendation and utterance generation.
It uses the user portrait to induce a reasoning tree based on the similarity between estimated user interests and nodes in the hierarchical knowledge measured using word embeddings.
Experimental results using ReDial dataset showed the proposed method recommends a wider variety of items and generates more diverse utterances than CR-Walker.
Our detailed analyses also suggested the proposed method can capture multiple user interests ranging from different classes.
Future work includes user studies using the proposed system rather than the investigation with given dialogue contexts of ReDial, pursuing sophisticated conversational recommendation strategies to narrow down user interests, and appropriate evaluation of entire CRS systems.

 \bibliographystyle{unsrt}
 \bibliography{anthology, referenc_my}

\backmatter
\appendix
%
%
%

\chapter{Chapter Heading}
\label{introA} 

Use the template \emph{appendix.tex} together with the Springer document class SVMono (monograph-type books) or SVMult (edited books) to style appendix of your book in the Springer layout.

\section{Section Heading}
\label{sec:A1}
Instead of simply listing headings of different levels we recommend to let every heading be followed by at least a short passage of text. Further on please use the \LaTeX\ automatism for all your cross-references and citations.

\subsection{Subsection Heading}
\label{sec:A2}
Instead of simply listing headings of different levels we recommend to let every heading be followed by at least a short passage of text. Further on please use the \LaTeX\ automatism for all your cross-references and citations as has already been described in Sect.~\ref{sec:A1}.

For multiline equations we recommend to use the \verb|eqnarray| environment.
\begin{eqnarray}
\vec{a}\times\vec{b}=\vec{c} \nonumber\\
\vec{a}\times\vec{b}=\vec{c}
\label{eq:A01}
\end{eqnarray}

\subsubsection{Subsubsection Heading}
Instead of simply listing headings of different levels we recommend to let every heading be followed by at least a short passage of text. Further on please use the \LaTeX\ automatism for all your cross-references and citations as has already been described in Sect.~\ref{sec:A2}.

Please note that the first line of text that follows a heading is not indented, whereas the first lines of all subsequent paragraphs are.

%
\begin{figure}[t]
\sidecaption[t]
\includegraphics[scale=.65]{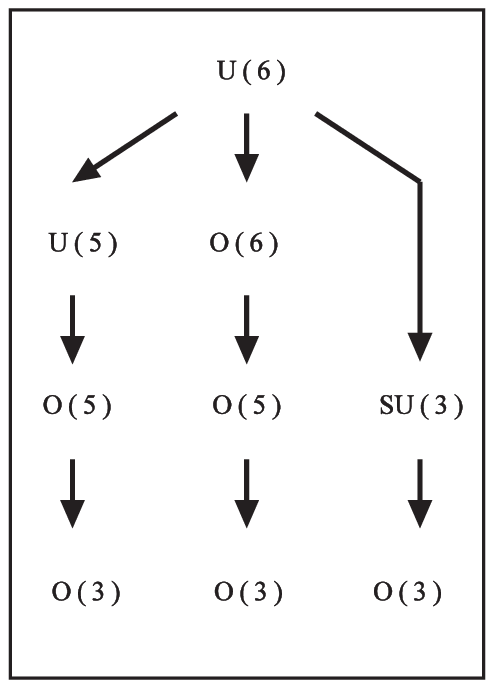}
%
%
\caption{Please write your figure caption here}
\label{fig:A1}       
\end{figure}

%
\begin{table}
\caption{Please write your table caption here}
\label{tab:A1}       
%
%
\begin{tabular}{p{2cm}p{2.4cm}p{2cm}p{4.9cm}}
\hline\noalign{\smallskip}
Classes & Subclass & Length & Action Mechanism  \\
\noalign{\smallskip}\hline\noalign{\smallskip}
Translation & mRNA$^a$  & 22 (19--25) & Translation repression, mRNA cleavage\\
Translation & mRNA cleavage & 21 & mRNA cleavage\\
Translation & mRNA  & 21--22 & mRNA cleavage\\
Translation & mRNA  & 24--26 & Histone and DNA Modification\\
\noalign{\smallskip}\hline\noalign{\smallskip}
\end{tabular}
$^a$ Table foot note (with superscript)
\end{table}
%

%
%

\Extrachap{Glossary}

Use the template \emph{glossary.tex} together with the Springer document class SVMono (monograph-type books) or SVMult (edited books) to style your glossary\index{glossary} in the Springer layout.

\runinhead{glossary term} Write here the description of the glossary term. Write here the description of the glossary term. Write here the description of the glossary term.

\runinhead{glossary term} Write here the description of the glossary term. Write here the description of the glossary term. Write here the description of the glossary term.

\runinhead{glossary term} Write here the description of the glossary term. Write here the description of the glossary term. Write here the description of the glossary term.

\runinhead{glossary term} Write here the description of the glossary term. Write here the description of the glossary term. Write here the description of the glossary term.

\runinhead{glossary term} Write here the description of the glossary term. Write here the description of the glossary term. Write here the description of the glossary term.
\printindex


\end{document}